\documentclass{article}

 \usepackage[preprint]{neurips_2025}

\usepackage[utf8]{inputenc} 
\usepackage[T1]{fontenc}    
\usepackage{hyperref}       
\usepackage{url}            
\usepackage{booktabs}       
\usepackage{amsfonts}       
\usepackage{nicefrac}       
\usepackage{microtype}      
\usepackage{xcolor}         
\usepackage{graphicx}
\usepackage{amsmath}
\usepackage{multirow}
\usepackage{threeparttable}
\usepackage{float}
\usepackage{adjustbox}
\usepackage{caption}
\usepackage{enumitem}

\newcommand{\AuthorVSpace}{1.5em}
\newcommand{\authsup}[1]{\,\textsuperscript{#1}}

\makeatletter

\def\AND{%
  \end{tabular}\par\vspace{\AuthorVSpace}%
  \noindent\makebox[\linewidth][c]{\begin{tabular}[t]{c}\bf\ignorespaces}%
}
\renewcommand{\@noticestring}{}
\makeatother

\title{Beyond the Strongest LLM: Multi-Turn Multi-Agent Orchestration vs. Single LLMs on Benchmarks}

\author{%
Aaron Xuxiang Tian\authsup{1}\thanks{Email: aarontian00@gmail.com} \quad
Ruofan Zhang\authsup{1} \quad
Jiayao Tang\authsup{2} \quad
Young Min Cho\authsup{3} \quad
\AND
Xueqian Li\authsup{4} \quad
Qiang Yi\authsup{4} \quad
Ji Wang\authsup{4} \quad
Zhunping Zhang\authsup{5} \quad
Danrui Qi\authsup{6} \quad
Zekun Li\authsup{7} \quad
\AND
Xingyu Xiang\authsup{8} \quad
Sharath Chandra Guntuku\authsup{3} \quad
Lyle Ungar\authsup{3} \quad
Tianyu Shi\authsup{4}\thanks{Co-corresponding author. Email: ty.shi@mail.utoronto.ca} \quad
\AND
Chi Wang\authsup{9}\thanks{Corresponding author. Email: chi@chiwang.cc}%
\\[\AuthorVSpace]
\parbox{\linewidth}{\centering
\authsup{1}Independent Researcher \quad
\authsup{2}Arizona State University \quad
\authsup{3}University of Pennsylvania\\
\authsup{4}Gradient \quad
\authsup{5}Nanyang~Technological~University \quad
\authsup{6}Simon Fraser University \quad
\authsup{7}University~of~California,~Santa~Barbara \quad
\authsup{8}Dropbox \quad
\authsup{9}Google~DeepMind
}
}

\begin{document}

\maketitle

\begin{abstract}
We study multi-turn multi-agent orchestration, where multiple large language model (LLM) agents interact over multiple turns by iteratively proposing answers or casting votes until reaching consensus. Using four LLMs (Gemini 2.5 Pro, GPT-5, Grok 4, and Claude Sonnet 4) on GPQA-Diamond, IFEval, and MuSR, we conduct two experiments: (i) benchmarking orchestration against single-LLM baselines; and (ii) ablations on GPQA-Diamond that vary whether agents see who authored answers and whether they can observe ongoing votes. Orchestration matches or exceeds the strongest single model and consistently outperforms the others. Analysis of best-achievable orchestration performance shows potential for further gains. The ablations show that revealing authorship increases self-voting and ties, and that showing ongoing votes amplifies herding, which speeds convergence but can sometimes yield premature consensus.
\end{abstract}

\section{Introduction}

No single LLM excels across benchmarks. \textbf{Multi-turn multi-agent orchestration} (hereafter referred to as \textbf{orchestration}), where multiple LLM agents coordinate over multiple turns, can combine complementary strengths, but outcomes depend heavily on how these turns are organized.
Prior work on multi-agent collaboration \citep{Kaesberg_2025, tran2025multiagentcollaborationmechanismssurvey, du2025multiagentcollaborationcrossteamorchestration} demonstrates potential but rarely compares with strong single-LLM baselines. Other studies investigate consensus mechanisms \citep{pitre-etal-2025-consensagent, Gao2025ASC, Yi2025FromDT} or orchestration frameworks \citep{Park2025MAPoRLMP, Chen2024BlockAgentsTB, Zhang2023ExploringCM}, yet none systematically examine how variations in coordination strategy affect multi-turn consensus. For example, it remains unclear whether revealing which agent produced which answer, or allowing agents to observe one another’s votes during voting, helps or harms the collective outcome.

We address this gap by evaluating an orchestration framework where heterogeneous LLM agents iteratively either propose new answers or cast votes over multiple turns to form consensus. Using four models (Gemini 2.5 Pro \citep{comanici2025gemini25pushingfrontier}, GPT-5 \citep{openai2025introducing_gpt5}, Grok 4 \citep{grok4_model_card_2025}, and Claude Sonnet 4 \citep{anthropic2025claude_sonnet4}), we conduct two complementary experiments:  
(1) Benchmark comparisons: orchestration vs.\ strong single-LLM baselines across GPQA-Diamond \citep{rein2023gpqa}, IFEval \citep{zhou2023instructionfollowingevaluationlargelanguage}, and MuSR \citep{sprague2024musrtestinglimitschainofthought}.  
(2) Coordination-strategy ablations: within orchestration on GPQA-Diamond, we vary whether agents know who wrote each answer and whether they can see votes that have already been cast to study how these factors affect turn-level behavior and consensus.

We preview three findings: (1) orchestration rivals the strongest single LLM across benchmarks, (2) simple coordination strategies measurably reshape turn-level behavior and consensus outcomes, and (3) the gap between best-achievable and actual orchestration performance shows clear headroom for better coordination.

Our main contributions are as follows: \\
\noindent $\bullet$ We systematically evaluate a multi-turn multi-agent orchestration framework, comparing it against strong single-LLM baselines across three benchmarks. \\
\noindent $\bullet$ We conduct controlled ablations to reveal how coordination strategies shape multi-turn dynamics. \\
\noindent $\bullet$ We analyze the gap between best-achievable and actual orchestration performance, exposing coordination failures despite some agents producing correct answers and highlighting opportunities for improved framework design.

\section{Methods}

We evaluate a multi-turn multi-agent orchestration framework that coordinates LLM agents via a structured protocol (see Appendix~\ref{app:framework} for the formal description).
Its coordination unfolds in 3 phases:
\begin{enumerate}
    \item \textbf{Agent Action}: agents operate asynchronously without waiting for others. 
    At each step, an agent may either generate a new candidate answer or cast a vote for an answer in the current answer set. 
    If a new answer is introduced \emph{during voting}, a \emph{dynamic restart} is triggered, interrupting the voting process so that all agents can reevaluate with the updated answer set.

    \item \textbf{Consensus}: once all agents have produced at least one answer and cast votes, the winning agent is selected as the one whose answer receives majority votes in the latest answer set.

    \item \textbf{Final Presentation}: the winning agent, selected by consensus, receives access to the \emph{latest answer set} produced during coordination, along with the associated voting feedback and reasoning. 
    It then synthesizes a comprehensive final answer that integrates insights from all participants, which is presented to the user as a single coherent response.
\end{enumerate}

\section{Experiments}
We conduct two experiments: (1) orchestration vs.\ single-LLM baselines, and (2) coordination-strategy ablations within orchestration.

Across both experiments, we use four LLMs (Gemini 2.5 Pro, GPT-5, Grok 4, and Claude Sonnet 4). 

\subsection{Baseline Comparisons}
\label{sec:baseline-comparison}

\paragraph{Goal.}
Evaluate how multi-agent orchestration performs compared to single-LLM baselines.

\paragraph{Setup.}
We compare two evaluation targets on each benchmark:  
(i) \textbf{Orchestration}: accuracy of the final answer produced by the multi-turn multi-agent orchestration framework,  
(ii) \textbf{LLM}: accuracy of each LLM when run independently.

\paragraph{Metrics.} \textbf{Accuracy}: proportion of tasks correctly solved under each evaluation target.

For statistical comparison, we use the exact McNemar’s test to assess paired differences between orchestration and each LLM. Results are reported in Appendix Table~\ref{tab:statistical_tests}.

\subsection{Coordination Strategy}
\label{sec:strategy-analysis}

\paragraph{Goal.}
Examine how different voting configurations affect consensus outcomes and voting behavior.

\paragraph{Setup.}
We vary two coordination variables that govern how agents vote:

\textbf{Voting identity disclosure} – controls whether the identity of the agent providing an answer is visible during voting.
\begin{itemize}[leftmargin=3em, nosep]
  \item \textbf{Anonymous Voting}: answers are attributed only to anonymized IDs (e.g., \texttt{agent1}, \texttt{agent2}), so voters cannot directly associate an answer with a specific model. 
  Since agents are memory-less and do not retain history across tasks, anonymization of names effectively guarantees that the voting process is identity-agnostic.
  \item \textbf{Identified Voting}: answers are attributed to the actual agent identity (e.g., \texttt{GPT-5}, \texttt{Claude}), allowing voters to see which model produced which answer.
\end{itemize}

\textbf{Vote tally visibility} – controls whether an agent can observe others’ votes before casting its own.
\begin{itemize}[leftmargin=3em, nosep]
  \item \textbf{Hidden Tally}: agents cannot observe any voting records before casting their own vote.
  \item \textbf{Visible Tally}: when voting, agents can see the votes cast so far for the current answer set, including (i) which agent voted for which answer and (ii) the reasons they provided. Displayed identities follow the chosen identity-disclosure setting (anonymized IDs or real model names).
\end{itemize}

The default setting is \textbf{Anonymous Voting + Hidden Tally}, and we vary one variable at a time.

\paragraph{Metrics.}
All rates are computed over all tasks in the benchmark.
\begin{itemize}
  \item \textbf{Self-Voting Rate}: proportion of votes where an agent voted for its own answer.
  \item \textbf{First-voted Selected Rate}: proportion of tasks where the agent whose answer received the first vote was ultimately selected as the winning agent by consensus.
  \item \textbf{Consensus Tie Rate}: proportion of tasks where the final consensus ended in a tie (no single answer received a majority of votes).
\end{itemize}

\subsection{Benchmarks}

We use three benchmarks: GPQA-Diamond, IFEval, and MuSR.  
Baseline comparisons (Section~\ref{sec:baseline-comparison}) use all three,  
while strategy analysis (Section~\ref{sec:strategy-analysis}) focuses on GPQA-Diamond.
See Appendix~\ref{app:benchmarks} for evaluation protocols.

\noindent $\bullet$ \textbf{GPQA-Diamond}: graduate-level; expert-curated; Google-proof questions requiring deep reasoning. \\
\noindent $\bullet$ \textbf{IFEval}: instruction-following tasks designed to test adherence to natural language instructions. \\
\noindent $\bullet$ \textbf{MuSR}: narrative reasoning tasks including murder mysteries and real scenarios.

\section{Results and Discussions}

\subsection{Performance across Benchmarks}

Table~\ref{tab:performance_comparison} compares orchestration against single-LLM baselines across three benchmarks. 
Orchestration achieves the highest accuracy on two of the three benchmarks and the best overall average.

\begin{table}[H]
\centering
\caption{Accuracy (\%) on three benchmarks. Each LLM row reports single-LLM accuracy. The "Orchestration" row reports a multi-turn multi-agent system that orchestrates four agents, each powered by Grok 4, GPT-5, Gemini 2.5 Pro, or Claude Sonnet 4. The last column reports the average accuracy across the three benchmarks. \textbf{Bold numbers} indicate the highest accuracy in each column.}
\label{tab:performance_comparison}
\begin{tabular}{l|c|c|c|c}
\toprule
Model & GPQA-Diamond & IFEval & MuSR & Avg. \\
\midrule
Grok 4          & 85.4 & 84.7 & 67.6 & 79.2 \\
GPT-5           & 84.8 & 87.4 & 69.2 & 80.5 \\
Gemini 2.5 Pro  & 85.9 & 66.0 & \textbf{69.6} & 73.8 \\
Claude Sonnet 4 & 68.2 & 63.6 & 62.8 & 64.9 \\
Orchestration   & \textbf{87.4} & \textbf{88.0} & 68.3 & \textbf{81.2} \\
\bottomrule
\end{tabular}
\end{table}

\paragraph{Orchestration rivals the strongest single LLM and clearly surpasses the weakest.}
Across all three benchmarks, orchestration matches or slightly exceeds the best single LLM, while outperforming the weakest model by a substantial margin. For instance, orchestration achieves 87.4\% on GPQA-Diamond (vs.\ 85.9\% for Gemini and 68.2\% for Claude), 88.0\% on IFEval (vs.\ 87.4\% for GPT-5 and 63.6\% for Claude), and 68.3\% on MuSR (vs.\ 69.6\% for Gemini and 62.8\% for Claude). The strongest LLM varies by dataset, yet orchestration consistently delivers top-tier accuracy without prior knowledge of which model is best.

\paragraph{Strong performance, with tangible potential for further gains.}
Across all three benchmarks, orchestration often fails despite at least one participating agent producing the correct answer, revealing untapped potential for better coordination. For instance, on GPQA-Diamond, at least one agent was correct in 95.5\% of cases, while orchestration reached only 87.4\%. Among the errors, 64\% of cases still had at least one correct agent, and in 31\% of these errors, two or more agents produced the correct answer, yet the consensus converged on an incorrect majority (Table~\ref{tab:benchmark_results}). These results indicate that orchestration already contains strong signals for success, and improved coordination mechanisms could better exploit this information and push performance closer to the best-achievable.

We examine two concrete examples showing that coordination can succeed when agents resolve discrepancies through self-correction, but can also fail when convincing yet incorrect analyses outweigh a correct answer. 
A detailed analysis of these cases is provided in Appendix~\ref{app:case_studies}.

\subsection{Coordination Strategy Analysis}

\begin{figure}[t]
    \centering
    \includegraphics[width=\linewidth]{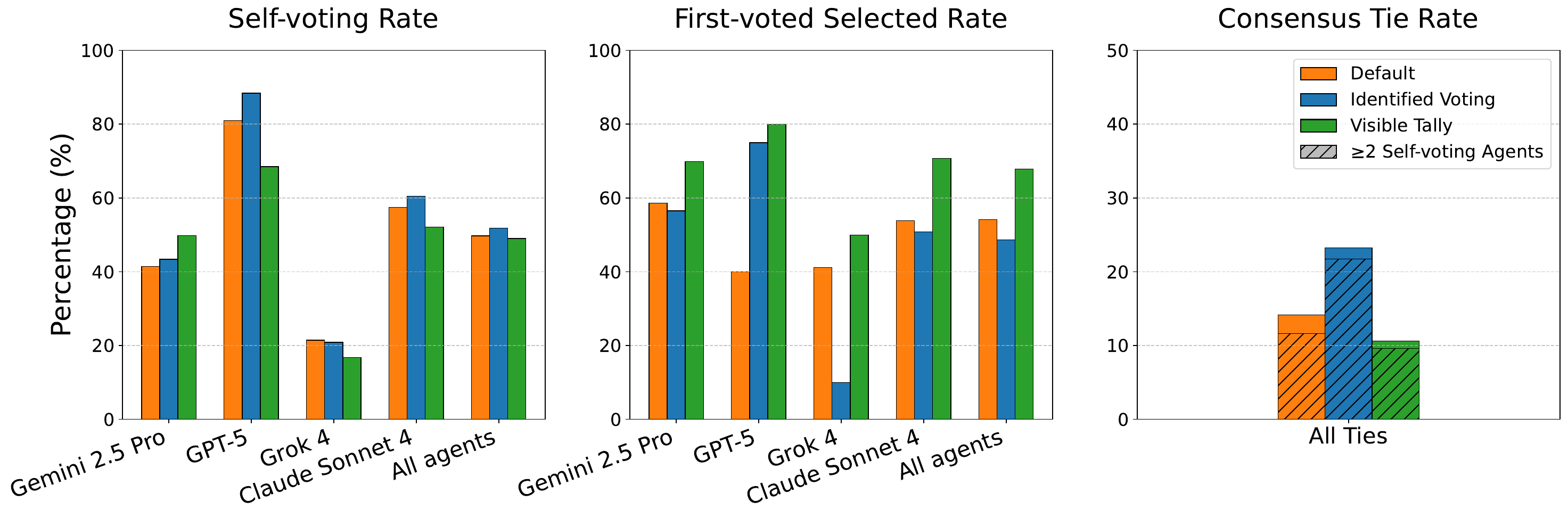}
    \caption{Effect of coordination strategies on GPQA-Diamond. Bars show percentages under three settings: Default (Anonymous + Hidden Tally), Identified Voting, and Visible Tally. Left: \textbf{Self-voting Rate}, the percentage of votes an agent cast for its own answer. Middle: \textbf{First-voted Selected Rate}, the percentage of tasks where the answer that received the first vote became the final consensus. In the first two plots, the rightmost group "All agents" aggregates across agents. Right: \textbf{Consensus Tie Rate}, the percentage of tasks with no majority. The hatched bar ("$\geq$2 Self-voters") marks the subset of tie cases where at least two agents voted for themselves. Models: Gemini 2.5 Pro, GPT-5, Grok 4, and Claude Sonnet 4.}
    \label{fig:strategy}
\end{figure}

We vary coordination strategies on GPQA-Diamond, with results shown in Figure~\ref{fig:strategy}.

\paragraph{Voting identity disclosure increases self-voting and biases consensus.}
Under Identified Voting, the \textbf{Self-Voting Rate} increased for three agents, with the largest jump for GPT-5 (81.0\%$\rightarrow$88.4\%). At the group level, GPT-5 became the consensus answer in 40.9\% of tasks (vs.\ 29.8\% under Anonymous Voting), while Claude and Grok-4 were chosen less often. The \textbf{Consensus Tie Rate} also rose sharply (14.1\%$\rightarrow$23.2\%). Analysis of tie cases shows that all 46 ties involved at least one self-voting agent, and 43 involved two or more—nearly doubling from 22 under Anonymous Voting. This confirms that revealing identities amplifies self-voting, skews consensus toward dominant agents, and makes agreement harder to reach.

\paragraph{Visible tallies amplify herding behavior.}
When agents could observe prior votes, the \textbf{First-voted Selected Rate} increased from 54.1\% to 67.8\% (+13.7\%). By agent, this effect was especially strong for GPT-5 (40.0\%$\rightarrow$80.0\%), but also present for Claude (+17.0\%), Gemini (+11.3\%), and Grok-4 (+8.8\%). Qualitative log inspection shows agents explicitly citing majority votes when justifying choices, reinforcing early momentum and drawing further votes to the majority. For example:  

\noindent $\bullet$ \emph{“has already received the majority of votes and provides comprehensive coverage of the topic.”} \\
\noindent $\bullet$ \emph{“The best current answer likely points to Agent~2 since it has the majority of votes. I need to use the vote tool to vote between Agent~2 and Agent~3, confirming it meets the original question.”}

\section{Conclusion}
\label{sec:conclusion}
We run two experiments with four LLMs on GPQA-Diamond, IFEval, and MuSR. Orchestration rivals or surpasses the strongest LLM. The gap between best-achievable and actual orchestration performance shows room for further gains. Coordination strategies strongly shape outcomes.

{\small
\bibliographystyle{plainnat}  
\bibliography{refs}
}

\appendix

\section{Detailed Benchmark Results}\label{app:detailed-results}

\begin{table}[H]
\centering
\caption{Exact McNemar’s test $p$-values comparing multi-agent orchestration with each single-LLM baseline on each benchmark. Values are reported in scientific notation with three significant figures; lower values indicate stronger evidence that accuracies differ.}
\label{tab:statistical_tests}
\begin{tabular}{lccc}
\toprule
Model & GPQA-Diamond & IFEval & MuSR \\
\midrule
Grok 4          & 4.81e-01 & 4.44e-02 & 5.18e-01 \\
GPT-5           & 3.59e-01 & 7.95e-01 & 9.34e-01 \\
Gemini 2.5 Pro  & 6.29e-01 & 1.08e-28 & 6.34e-01 \\
Claude Sonnet 4 & 1.38e-07 & 5.72e-33 & 3.25e-03 \\
\bottomrule
\end{tabular}
\end{table}

\begin{table}[H]
\centering
\caption{Coverage of correct answers by participating agents under the multi-agent orchestration framework across benchmarks. 
The first column reports the percentage of tasks where at least one agent produced the correct answer. 
The second column reports, relative to those tasks, the percentage where two or more agents were correct but the consensus outcome was still incorrect.}
\label{tab:benchmark_results}

\begin{adjustbox}{max width=\textwidth}
\begin{tabular}{l|cc}
\toprule
Benchmark & $\geq$1 agent correct but wrong consensus (\%) & $\geq$2 agents correct (subset of col.\ 1) (\%) \\
\midrule
GPQA-Diamond   & 64 & 31 \\
IFEval & 62 & 55 \\
MuSR   & 64 & 53 \\\bottomrule
\end{tabular}
\end{adjustbox}

\end{table}

\section{Case Studies: Detailed Analysis}
\label{app:case_studies}

\subsection{Success Case: Self-Correction Through Peer Observation}

\paragraph{Question.}
Observations of a quasar across the electromagnetic spectrum are being carried out. 
Focusing on the near infrared and optical spectrum, there is a peak at wavelength of about 790 nm, 
and at shorter wavelengths < 790 nm the flux drops significantly. 
If we lived in a universe defined by Lambda-CDM model such that the current Hubble constant 
is 70 km s$^{-1}$ Mpc$^{-1}$, matter density parameter is 0.3, dark energy density parameter is 0.7, 
and universe is flat, what can be assumed about the value of the comoving distance 
(for scale factor $a=1$) of this quasar from the Earth? \\
A. 8 Gpc \quad B. 7 Gpc \quad C. 6 Gpc \quad D. 9 Gpc

\paragraph{Correct Answer.}
A. 8 Gpc

\paragraph{Collaboration Process.}
The Claude Sonnet 4 agent initially calculated a comoving distance of 6 Gpc. 
Upon observing discrepancies among the other agents' answers, it noted: 

\begin{quote}
\emph{There is a significant discrepancy in their comoving distance calculations: 
Agent1 calculates $\sim$6.1 Gpc and chooses C (6 Gpc), 
Agent2 calculates $\sim$8.95 Gpc and chooses D (9 Gpc).}
\end{quote}

It then re-examined their approaches, performed multiple verification steps, 
and revised its own answer to 8 Gpc: 

\begin{quote}
\emph{For a flat Lambda-CDM model with $z=5.5$ and the given cosmological parameters, 
standard cosmological calculators and numerical integration yield a comoving distance 
of approximately 8000–8500 Mpc, which equals 8.0–8.5 Gpc. 
Among the given options, this is closest to 8 Gpc.}
\end{quote}

With this revised and well-supported reasoning, three of the four agents 
selected option A, leading the system to converge on the correct answer.

\paragraph{Success Mechanism.}
This case shows how peer observation of discrepancies prompted re-evaluation 
and self-correction, enabling the orchestration system to converge on the correct answer.

\subsection{Limitation Case: Analysis Quality Over Answer Correctness}

\paragraph{Question.}
How many of the stars listed below would be detectable using the ESPRESSO spectrograph, 
when it is coupled with one of the 8m VLT telescopes at the Paranal Observatory? 
A star is considered detectable if a signal-to-noise ratio (S/N) of at least 10 per binned pixel 
during a 1-hour exposure is achieved.

For more details about the ESPRESSO spectrograph, please refer to the following link: 
\url{https://www.eso.org/sci/facilities/paranal/instruments/espresso/overview.html}

a) Canopus \\
b) Polaris \\
c) Star with RA = 0 deg and DEC = 0 deg, Absolute V magnitude of 15 mag and located at 10 pc distance from us. \\
d) Star with RA = 0 deg and DEC = 0 deg, Absolute V magnitude of 15 mag and located at 200 pc distance from us. \\
e) Star with RA = 0 deg and DEC = 0 deg, Absolute V magnitude of 15 mag and located at 5 pc distance from us. \\
f) Star with RA = 0 deg and DEC = 0 deg, Absolute V magnitude of 15 mag and located at 50 pc distance from us. \\

A. 2 \quad B. 3 \quad C. 5 \quad D. 4

\paragraph{Correct Answer.}
A. 2

\paragraph{Collaboration Process.}
The Claude Sonnet 4 agent correctly identified that Canopus would saturate the detector 
in a 1-hour exposure, leading to the correct answer of 2 stars. 
However, the other three agents failed to consider detector saturation effects 
and incorrectly concluded that 3 stars were detectable.  

Despite Claude Sonnet 4 providing the correct reasoning, the orchestration system instead selected Gemini 2.5 Pro, noting:  

\begin{quote}
\emph{Agent 2 provides the most accurate and comprehensive reasoning. 
It correctly identifies that Polaris is not visible, that two stars are too faint, 
and most importantly, that Canopus is too bright and would saturate the detector.}
\end{quote}

This shows that the orchestration prioritized perceived reasoning quality over actual correctness, 
leading to the wrong final answer (3 stars).

\paragraph{Limitation Mechanism.}
This case illustrates how the system can be misled by detailed but incorrect analysis, 
underscoring the need to better balance correctness and reasoning quality.

\section{Benchmarks and Evaluation Protocols}
\label{app:benchmarks}

We use three benchmarks: GPQA-Diamond, IFEval, and MuSR. 
Baseline comparisons (Section~\ref{sec:baseline-comparison}) use all three, 
while strategy analysis (Section~\ref{sec:strategy-analysis}) focuses on GPQA-Diamond. 
Below we provide more detailed descriptions and the evaluation protocols for each.

\begin{itemize}
  \item \textbf{GPQA-Diamond}: graduate-level, expert-curated, Google-proof questions in 
  biology, physics, and chemistry requiring deep reasoning.  
  \hspace{1em}\textbf{Evaluation protocol:} multiple-choice; scored as correct/incorrect 
  per selected option.

  \item \textbf{IFEval}: instruction-following tasks across diverse categories designed to 
  test adherence to natural language instructions.  
  \hspace{1em}\textbf{Evaluation protocol:} evaluated using 
  \textit{Inst-level loose-accuracy} as defined in the benchmark.

  \item \textbf{MuSR}: narrative reasoning tasks, including murder mysteries and 
  real-world scenarios, generated via neurosymbolic methods.  
  \hspace{1em}\textbf{Evaluation protocol:} multiple-choice on Hugging Face; 
  scored by matching predicted and gold answer index.
\end{itemize}

\section{Formal Framework Description}\label{app:framework}

This appendix provides a detailed description of the multi-agent orchestration framework, which coordinates multiple language model agents through an exclusive action constraint. The framework provides a structured mechanism for organizing heterogeneous agents and serves as the basis of our experimental analysis. The orchestration unfolds in three coordination phases: (1) Agent Action, (2) Consensus, and (3) Final Presentation.

\subsection{State Model}

The framework is managed by an \textbf{Orchestrator}, which controls a set of heterogeneous agents:
\[
\mathcal{A} = \{a_1, a_2, \ldots, a_n\}.
\]

Each agent $a_i$ has an associated state $s_i \in \mathcal{S}$ defined as:
\[
s_i = \langle \text{answer}_i, \text{has\_voted}_i, \text{votes}_i, \text{restart\_pending}_i, \text{is\_killed}_i, \text{timeout\_reason}_i \rangle.
\]

The orchestrator maintains a global coordination state:
\[
\mathcal{S}_{\text{orchestrator}} = \langle \text{workflow\_phase}, \text{current\_task}, \text{selected\_agent}, \text{total\_tokens}, \text{coordination\_start\_time} \rangle.
\]

\subsection{Exclusive Action Constraint}

At each step, every agent must choose exactly one of two mutually exclusive actions:

\begin{enumerate}
    \item \textbf{Vote Action}: $V(a_i, a_j, r)$ where agent $a_i$ votes for $a_j$'s answer with reason $r$.
    \item \textbf{Answer Generation Action}: $N(a_i, c)$ where agent $a_i$ provides new content $c$.
\end{enumerate}

This exclusivity constraint is enforced as:
\[
\forall a_i \in \mathcal{A}: \text{action}(a_i) \in \{V, N\}, \quad |\text{action}(a_i)| = 1.
\]

The framework prevents mixed actions through strict enforcement:
\[
\resizebox{\textwidth}{!}{$
\text{valid\_response}(a_i) \iff 
\left(|\{V(a_i, \cdot, \cdot)\}| = 1 \land |\{N(a_i, \cdot)\}| = 0\right) 
\lor \left(|\{V(a_i, \cdot, \cdot)\}| = 0 \land |\{N(a_i, \cdot)\}| = 1\right)
$}
\]

\subsection{Coordination Workflow}

The framework operates over a set of heterogeneous agents. Its coordination unfolds in three phases with dynamic restart capabilities:

\paragraph{Phase 1: Agent Action.}
Agents operate asynchronously without waiting for others. At each step, an agent may either:
\begin{itemize}
    \item generate a new candidate answer $c_i$ using the answer generation tool, or
    \item cast a vote for an answer in the current answer set using the voting tool.
\end{itemize}

If a new answer is introduced \emph{during voting}, a \emph{dynamic restart} is triggered:
\[
\text{reset\_signal} = \text{True} \iff \exists a_i \in \mathcal{A}: N(a_i, c_i) \text{ is executed}.
\]

When reset occurs, all other agents are marked for restart, interrupting the voting process:
\[
\forall a_j \in \mathcal{A} \setminus \{a_i\}: s_j.\text{restart\_pending} = \text{True}.
\]

This ensures all agents can reevaluate with the updated answer set, preventing premature consensus. The restart mechanism preserves the latest answer set:
\[
\mathcal{C}_{\text{latest}} = \{c_i : s_i.\text{answer} \neq \emptyset \land s_i.\text{is\_killed} = \text{False}\}.
\]

\paragraph{Phase 2: Consensus.}
Once all agents have produced at least one answer and cast their votes, the winning agent is selected as the one whose answer receives the majority of votes in the latest answer set. If a dynamic restart is triggered, all prior votes are discarded and only votes cast after the restart are considered.

\[
\text{winner} = \arg\max_{a_i \in \mathcal{A}} \sum_{a_j \in \mathcal{A}} \mathbb{I}[V(a_j, a_i, \cdot) \land \neg s_j.\text{restart\_pending}],
\]

where $\mathbb{I}[\cdot]$ is the indicator function and votes from agents with pending restarts are invalidated. Ties are broken by the order in which agents are configured in the system.

\paragraph{Phase 3: Final Presentation.}
The winning agent, selected by consensus, receives access to the \emph{latest answer set} produced during coordination, along with the associated voting feedback and reasoning. It then synthesizes a comprehensive final answer that integrates insights from all participants, which is presented to the user as a single coherent response.

When vote tallies are visible, the displayed information (which answers have received votes from which agents, and their reasons) 
respects the chosen identity-disclosure setting (anonymous IDs or real model names).

\subsection{Workflow Tools and Enforcement}

The framework provides two mandatory workflow tools that agents must use:

\begin{enumerate}
    \item \textbf{Voting Tool}: $\texttt{vote}(\text{agent\_id}, \text{reason})$
    \begin{itemize}
        \item Requires existing answers in the current answer set
        \item Invalid when $\mathcal{C}_{\text{latest}} = \emptyset$
        \item The visibility and form of agent identities follow the coordination strategies:
(i) under \emph{Anonymous Voting}, IDs are anonymized as \(\{\texttt{agent\_1}, \texttt{agent\_2}, \ldots\}\),
(ii) under \emph{Identified Voting}, real agent names are shown like \(\{\texttt{GPT5-5}, \texttt{Grok 4}, \ldots\}\).
    \end{itemize}
    
    \item \textbf{Answer Generation Tool}: $\texttt{new\_answer}(\text{content})$
    \begin{itemize}
        \item Triggers dynamic restart for all other agents
        \item Content must be unique (no duplicate answers allowed)
        \item Resets voting state: $\forall a_i \in \mathcal{A}: s_i.\text{has\_voted} = \text{False}$
    \end{itemize}
\end{enumerate}

Tool enforcement ensures compliance:
\[
\text{workflow\_compliant}(a_i) \iff \exists t \in \{\texttt{vote}, \texttt{new\_answer}\}: t \text{ is called exactly once}.
\]


\end{document}